\documentclass[letterpaper, 10 pt, conference]{ieeeconf}  % Comment this line out if you need a4paper

\usepackage{graphicx}% remove the 'demo' option in the real document
\usepackage{enumerate}% http://ctan.org/pkg/enumerate
\usepackage{amsmath}
\usepackage{algorithm}
\usepackage[noend]{algpseudocode}
\usepackage{subfigure}
\usepackage{chngcntr}
\usepackage{graphicx}
\usepackage{gensymb}

\makeatletter
\let\NAT@parse\undefined
\makeatother
\usepackage[table,xcdraw]{xcolor}

\usepackage[sort&compress,numbers]{natbib}

\usepackage{url}
\usepackage{tikz}

\usepackage{caption}
\usepackage{hyperref}
\hypersetup{
	colorlinks = true
}

\usepackage[colorinlistoftodos]{todonotes}

\usepackage[font=footnotesize,labelfont=bf,tableposition=top]{caption}

\usepackage{etoolbox}
\makeatletter
\patchcmd{\@makecaption}
{\scshape}
{}
{}
{}
\makeatletter
\patchcmd{\@makecaption}
{\\}
{.\ }
{}
{}
\makeatother

\usepackage{color}

\IEEEoverridecommandlockouts

\usepackage{caption}
\usepackage[singlelinecheck=false, font=footnotesize]{caption}
\usepackage{etoolbox}
\makeatletter
\patchcmd{\@makecaption}
{\scshape}
{}
{}
{}
\makeatletter
\patchcmd{\@makecaption}
{\\}
{.\ }
{}
{}
\makeatother

\usepackage{amsfonts}
\usepackage{amssymb}

\makeatletter
\def\BState{\State\hskip-\ALG@thistlm}
\makeatother

\usepackage{epstopdf}
\usepackage{epsfig}
\usepackage{multirow}
\usepackage{soul}

\begin{document}
	\title{\Large \bf SegICP:  Integrated Deep Semantic Segmentation and Pose Estimation}
	
	\author{Jay M. Wong, Vincent Kee${}^{\dagger}$, Tiffany Le${}^{\dagger}$, Syler Wagner, Gian-Luca Mariottini, \\Abraham Schneider, Lei Hamilton, Rahul Chipalkatty, Mitchell Hebert, and David M.S. Johnson\\
		\emph{Draper, Cambridge, MA, USA } \\\mbox{}\\
		Jimmy Wu, Bolei Zhou, and Antonio Torralba\\
		\emph{Massachusetts Institute of Technology, Cambridge, MA, USA}
		\thanks{Draper Fellows${}^\dagger$ associated with both Draper and MIT. Any opinions, findings, conclusions, or recommendations expressed in this material are solely those of the author(s) and do not necessarily reflect the views of these organizations. We thank Leslie Pack Kaelbling, Tom{\'a}s Lozano-P{\'e}rez, Scott Kuindersma, and Russ Tedrake for their insight and feedback. \textbf{Corresponding author:} Jay M. Wong {\href{mailto:jmwong@draper.com}{\texttt{jmwong@draper.com}}}}
	}
	
	\maketitle

	\IEEEpeerreviewmaketitle
	
	\begin{abstract}
		Recent robotic manipulation competitions have highlighted that  sophisticated robots still struggle to achieve fast and reliable perception of task-relevant objects in complex, realistic scenarios. To improve these systems' perceptive speed and robustness, we present SegICP, a novel integrated solution to object recognition and pose estimation. SegICP couples convolutional neural networks and multi-hypothesis point cloud registration to achieve both robust pixel-wise semantic segmentation as well as accurate and real-time 6-DOF pose estimation for relevant objects.%, even in the presence of occlusions and sensor noise. 
		Our architecture achieves $1~\text{cm}$ position error and $<5^\circ$ angle error in real time \emph{without} an initial seed. We evaluate and benchmark SegICP against an annotated dataset generated by motion capture. 
	\end{abstract}
	
	\section{Introduction}
	To achieve robust, autonomous operation in unstructured environments, robots must be able to identify relevant objects and features in their surroundings, recognize the context of the situation, and plan their motions and interactions accordingly. Recent efforts in autonomous manipulation challenges such as the DARPA Robotics Challenge~\cite{pratt2013darpa} and the Amazon Picking Challenge~\cite{wurman2015amazonpicking} resulted in state-of-the-art perception capabilities enabling systems to perceive, reason about, and manipulate their surroundings. However, existing object identification and pose estimation solutions for closed-loop manipulation tasks are generally (1) not robust in cluttered environments with partial occlusions, (2) not able to operate in real-time (${<}1~\text{Hz}$), (3) not sufficiently accurate~\cite{zeng2016multi}, or (4) incapable of high accuracy without good initial seeds \cite{schmidt2014dart}.
	
\begin{figure}[t]
		\centering
		%\vspace{+30mm}
		\mbox{
			\hspace{-7.5pt}
			\subfigure{\includegraphics[width=1.65in]{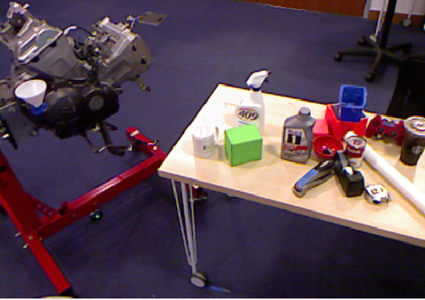} }
			\hspace{-7.5pt}
			\subfigure{\includegraphics[width=1.65in]{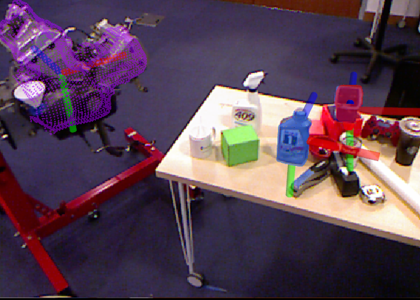} }
		}
		\vspace{-15pt}
		\caption{Given an RGB image (left) and depth frame, our SegICP approach segments the objects in a pixel-wise fashion and estimates the 6 DOF pose of each object with $1~\text{cm}$ position and $5^\circ$ angle error (right).}
		\label{intro-fig}
	\end{figure}

	We present a novel perception pipeline that tightly integrates deep semantic segmentation and model-based object pose estimation, achieving real-time pose estimates with a median pose error of $1~\text{cm}$ and ${<}\, 5^\circ$. Our solution (referred to as \emph{SegICP}) uses RGB-D sensors (and proprioceptive information when available) to provide semantic segmentation of all relevant objects in the scene along with their respective poses (see Figure~\ref{intro-fig}) in a \emph{highly parallelized} architecture. 
	
	The main contributions of this manuscript are as follows:
	\begin{enumerate}
		\item A \emph{highly parallelized} approach to integrated semantic segmentation and multi-hypothesis object pose estimation with $1~\text{cm}$ accuracy with a single view operating at $70$--$270~\text{ms}$ ($4$--$14$ Hz) \emph{without any prior pose seeds}.
		\item A novel metric to score the quality of point cloud registration, allowing for autonomous and accurate pose initialization over many potential hypotheses.
		\item An efficient automatic data-collection framework for acquiring annotated semantic segmentation and pose data by using a motion capture system. 
		\item Analysis and benchmarking of our SegICP pipeline against the automatically annotated object poses. 
	\end{enumerate}
	
	\begin{figure*}
		\hspace{-27.5pt}
		\includegraphics[width=1.0725\textwidth]{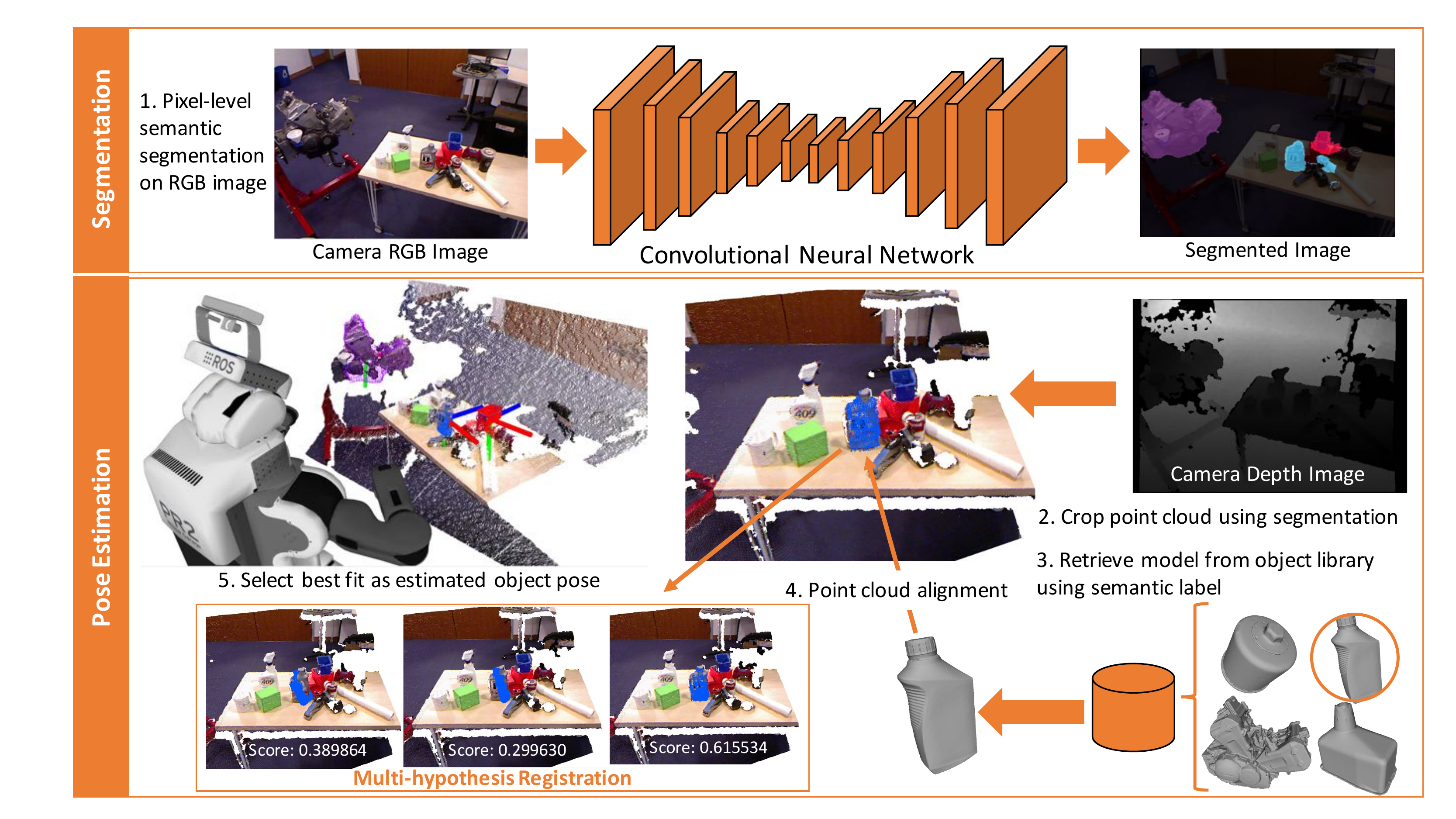}
		\vspace{-20pt}%
		\caption{\textbf{The full SegICP pipeline operating in a cluttered environment.} The system detects objects relevant to the automotive oil change task and estimates a 6-DOF pose for each object. The colored overlay pixels in the segmented image (top right) correspond to a blue funnel (red), an oil bottle (blue), and the engine (purple), as detected by the Kinect1 mounted on top of a PR2 robot. Selected multi-hypothesis registrations for the oil bottle object (bottom left) are shown with their respective alignment scores. The hypothesis registrations are evaluated in parallel to determine the optimal object pose.}
		\label{segicp-pipeline}
	\end{figure*}	
	
	\section{Related Work}
	Our approach builds on the substantial literature devoted to robot perception of mobile manipulation task environments and the relevant objects therein. Robot systems must be able to first identify entities that pertain to the task at hand and reason about their relative poses to eventually manipulate and interact with them. Accordingly, we discuss the relevant literature in object recognition and pose estimation. 
	
	\textbf{Object Recognition.} \; \; 
	Semantic segmentation, which assigns each pixel in an image to one of a set of predefined categories, effectively solves the object recognition problem. This approach is in contrast to that of many object recognition systems, which only output bounding boxes around objects of interest \cite{girshick2014rich,zhou2016fast,ren2015faster}. Although the winning team for the 2015 Amazon Picking Challenge used a bag-of-words approach \cite{Liu2016} instead of per-pixel categorization, there are multiple advantages to retaining the spatial position of every object. Particularly in robotic applications such as grasping or autonomous driving, semantic segmentation enables a higher resolution representation of the arrangement, identities of objects in a cluttered scene, and effectively addresses self-occlusions. Previous approaches to object recognition have used classifiers with hand-engineered features~\cite{shotton2008semantic,brostow2008segmentation} to generate either bounding-box object locations or noisy pixel-wise class predictions, which are then smoothed using CRFs~\cite{ladicky2010and}. Recent work in computer vision has shown that convolutional neural networks (CNNs) considerably improve image classification~\cite{lecun1998gradient,krizhevsky2012imagenet}; CNNs originally trained for image classification can be successfully re-purposed for dense pixel-wise semantic segmentation~\cite{Badrinarayanan2015,noh2015learning,long2015fully,zhou2016semantic,yu2015multi}. Such approaches generally retain the lower level feature detectors of image classification models such as AlexNet~\cite{krizhevsky2012imagenet} or VGG \cite{Simonyan2014}  and stack on additional layers, such as convolution \cite{Badrinarayanan2015}, deconvolution \cite{noh2015learning}, or dilated convolution \cite{yu2015multi}.% for dense prediction.
	
	\textbf{Pose Estimation.} \;\; 
	While semantic segmentation is able to identify and locate objects in 2D images, pose estimation refines object location by also estimating the most likely 6-DOF pose of each identified object.
	Previous approaches to this task have used template matching, which can recover the pose of highly-textured objects \cite{huttenlocher1993comparing} \cite{lowe2001local} using local features such as SIFT ~\cite{lowe2004distinctive}. For RGB-D images, use of stable gradient and normal features has been demonstrated with LINEMOD \cite{hinterstoisser2012model,rios2013discriminatively}. Approaches using parts-based models have also been successful~\cite{savarese20073d,pepik2012teaching,lim2014fpm}. However, these methods are not robust to variations in illumination or to scene clutter~\cite{zeng2016multi}.  While a class of point cloud registration algorithms attempt to solve the global optimization problem \cite{zhou2016fast}, such approaches rely on surface normals features and degrade when objects are generally flat, have low quantities of informative features, or exhibit potentially ambiguous geometries.  
	
	A widely accepted approach to pose estimation is the class of iterative closest point (ICP) registration algorithms~\cite{besl1992method,rusinkiewicz2001efficient,yuan20163d}. These approaches usually require initialization close to the global optima as gradient-based methods tend to fall into poor local minima and are not robust to partial or full occlusions \cite{schmidt2014dart}. Most relevant to our work, Team MIT-Princeton demonstrated promising results in the Amazon Picking Challenge using multiple views with a fully convolutional neural network to segment images and fit 3D object models to the segmented point cloud \cite{zeng2016multi}. However, their pose estimation system was slow (${\sim}1~\text{s}$ per object) and showed high position and angle errors ($5~\text{cm}$ and ${\sim}15^\circ$).
	We advance this prior work by presenting a novel metric for scoring model registration quality, allowing accurate initial pose estimation through multi-hypothesis registration. %From a single RGB-D frame, we achieve increased pose-tracking robustness and accuracy. 
	Further, we emphasize an order of magnitude speedup by leveraging a \emph{highly parallelized} design that operates over all objects simultaneously. We couple these advances with an efficient data collection pipeline that automatically annotates semantic segmentation labels and poses for relevant objects. %Overall, SegICP achieves estimation errors on the order of $5^\circ$ and $1~\text{cm}$ and an overall run-time of $70$--$270~\text{ms}$ per object, \emph{without the need of an initial seed}.
	
	\section{Technical Approach}
	We present SegICP, a novel perceptual architecture that handles sensor input in the form of RGB-D and provides a semantic label for each object in the scene along with its associated pose relative to the sensor. SegICP acquires and tracks the 6-DOF pose of each detected object, operating at ${\sim}70~\text{ms}$ per frame ($270~\text{ms}$ during initialization phase) with $1~\text{cm}$ position error and ${<}\,5^\circ$ angle error, and can robustly deal with prolonged occlusions and potential outliers in the segmentation with a Kalman filter. SegICP achieves this using an object library approach to perception, referencing scanned 3D models of known objects, and performs 3D point cloud matching against cropped versions of these mesh models. %An example of object meshes pertaining to our automotive domain is illustrated in Figure~\ref{segicp-pipeline}. 
	In our architecture, as outlined in Figure~\ref{segicp-pipeline}, RGB frames are first passed through a CNN which outputs a segmented mask with pixel-wise semantic object labels. This mask is then used to crop the corresponding point cloud, generating individual point clouds for each detected object. ICP is used to register each object's point cloud with its full point cloud database model and estimate the pose of the object with respect to the sensor.
	
	\subsection{Semantic Segmentation by Neural Networks}
	Contrary to classical segmentation problems, we are specifically concerned with generating \emph{appropriate} masks over the depth map to aid accurate pose estimation. In an attempt to address this, we experimented with various CNN architectures that semantically segment known objects of interest.
	We explored two different CNN architectures, SegNet \cite{Badrinarayanan2015} and DilatedNet \cite{yu2015multi} (further discussed and elaborated in Section~\ref{compare-segs}). Of the two networks, we found that the best model for our SegICP pipeline was SegNet, a $27$-layer, fully convolutional neural network with $30$ million parameters. %based off SegNet \cite{Badrinarayanan2015} and VGGNet \cite{Simonyan2014}. 
	The network was trained on cropped and downsampled images from the training set (to 320$\times$320 pixels) consisting of eight object classes (including background) using the cross entropy criterion coupled with data augmentation consisting of image rotations, crops, horizontal and vertical flips, and color and position jitter. We further elaborate on the acquisition of annotated training data in Section~\ref{sec:data-collection}.
	
	\begin{figure}
		\centering
		\includegraphics[width=0.485\textwidth]{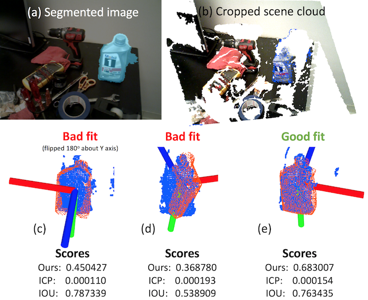}%
		\vspace{-1pt}
		\caption{\textbf{Examples of Multi-Hypothesis Registration Ranking:} The segmentation generated in (a) is used to produce the cropped scene cloud highlighted in blue (b). Panels (c-e) illustrates the registration of various candidate model crops (orange) to the cropped scene cloud (blue), along with their respective alignment scores.}%
		\vspace{-10pt}
		\label{multi-examples}
	\end{figure}    
	
	\subsection{Multi-Hypothesis Object Pose Estimation} 
	The resulting segmentation is used to extract each object's 3D point cloud from the scene cloud. The identity of each segmented object (the object's semantic label) predicted by SegNet is then used to retrieve its corresponding 3D mesh model from the object model library. The mesh model is converted into a point cloud representation, downsampled, and registered against its respective segmented point cloud.
	
	Point cloud registration is divided into two phases: \emph{acquisition} and \emph{tracking}. The objective of the acquisition phase is to find the initial optimal alignment between each object's model and its corresponding scene point cloud. This alignment is used to determine the visible side of the model (\emph{model crop}) and to initialize the tracking phase, whose objective is to fuse camera and robot motion information to maintain an accurate, real-time pose estimate of the object even during camera motion and potential occlusions. We use a point-to-point ICP~\cite{pcl} algorithm for registration. A contribution of this paper is the model-to-scene alignment metric that is used to determine the registration quality as well as switching between acquisition and tracking phases. % to decide when pose tracking has failed and to then switch back to acquisition.
	
	\textbf{The Acquisition Phase.}\,\,	The acquisition phase finds the initial optimal alignment and crop of the object's mesh model with the current point cloud. Multiple candidate crops are obtained by rendering the visible object's model at various azimuth and elevation angles and cropping the model to keep only the front face. Each of the candidate crops is initialized at the median position of the object's scene point cloud in order to remove segmentation outliers and prevent ICP from converging to incorrect local minima. In parallel, each candidate crop is run through a few iterations of the tracking phase to achieve a pose hypothesis. 
	
	A novel model-to-scene alignment metric is evaluated on each candidate model crop. The motivation behind the metric is to determine whether a candidate cloud can align well with the object cloud by finding the number of points in the candidate cloud with a unique corresponding match in the object's cloud. Letting $\mathcal{M}_i$ be the set of points in the candidate crop point cloud and $S$ be the set of points in the segmented object scene cloud, the alignment metric is given by:
	%\begin{equation}
	$\text{a}(\mathcal{M}_i, S) = \frac{|c|}{|\mathcal{M}_i|}$
	%\label{eqn:nn_metric}
	%\end{equation}
	where $c$ is the set of points in $\mathcal{M}_i$ with unique corresponding points in $S$ at most $\tau$ meters away. To compute the metric, SegICP first builds a kd-tree with $S$ and perform radius searches with a radius of $\tau$ meters from every point in $\mathcal{M}_i$. Each point in $\mathcal{M}_i$ is mapped to the closest  point in $S$ within $\tau$ that has not been already mapped to another point in $\mathcal{M}_i$, and then is added to $c$. 
	
	Figure~\ref{multi-examples} show examples of model crops and their respective scores. In particular, we illustrate metrics such as the ICP fitness score (a Euclidean error score) and intersection over union (IOU)\footnote{IOU is computed between the predicted segmentation and the projection of the registered model into the camera optical frame.} do not effectively distinguish good registrations from erroneous ones. In comparison, our proposed metric addresses these immediate shortcomings present on objects with high degrees of symmetry (e.g. the oil bottle). If any candidate scores are above a threshold $\epsilon$, SegICP switches to the tracking phase for future frames. % using highest-scoring fit to initialize tracking. The candidate with the best metric score is then used to initialize the tracking pipeline.
	
	% Removed this def b/c technically not correct and not as clear
	% $c\triangleq\{p\in\mathcal{M}_i: \Vert p - NN(p, S)\Vert \leq \tau\}$
	% 	The acquisition phase finds the initial optimal alignment and crop of the object's mesh model with the current point cloud. Multiple candidate crops are obtained by rendering the visible object's model at various azimuth and elevation angles and cropping the model to extract the model's front face. Each of the candidates is initialized at the median point position of the object's scene point cloud to remove segmentation outliers and prevent ICP from getting stuck in incorrect local minima. A few iterations of the tracking phase are run with each of the candidates in a multi-threaded fashion to obtain a pose hypothesis for each potential model crop.
	
	% 	A novel model-to-scene alignment metric (Eq. \ref{eqn:nn_metric}) is evaluated with each of the candidate model crops. If any candidate scores are above a threshold $\epsilon$, SegICP switches to the tracking phase for future frames. The candidate with the best metric score is used to initialize the tracking pipeline. The idea behind the alignment score is to find how many points in the object's model have a potentially successful match with the segmented object's scene cloud. We do this by using Nearest Neighbors (NN) between the number of points in the model candidate $\mathcal{M}_i$ that have a match in the segmented scene point cloud $S$, and evaluating their proportion on the overall number of points in  $\mathcal{M}_i$:

	\textbf{The Tracking Phase.}\,\,
	The candidate model pose and crop with the highest alignment score are used to initialize the tracking phase. In order to make the tracking procedure robust to imperfections on the boundary of the object's segmentation, the object's scene point cloud is further pruned by removing points outside a bounding box of the latest registered model pose. The pose obtained by registration is used as a measurement update in a Kalman filter to track each object's 6-DoF pose and velocities. By fusing known camera motions from the available odometry of the robot, the filter is able to handle temporary object occlusions and outlier pose estimates. Our alignment metric is evaluated on the fit to measure the uncertainty of the current pose measurement and to inform the Kalman filter accordingly. If the score goes below a minimum threshold $\theta$, the Kalman filter propagates the objects' pose based on odometry (and until a maximum pose uncertainty) while switching back to acquisition mode.
	
	\subsection{Automatically Annotating Training Data}
	\label{sec:data-collection}
	\begin{figure}
		\centering
		\includegraphics[width=0.485\textwidth]{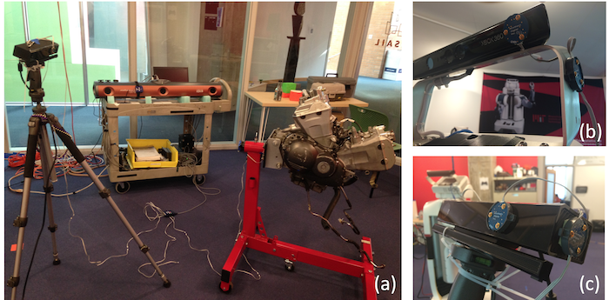}
		\caption{\textbf{Motion Capture System:} Setup using the NDI 3D Investigator Motion Capture System (a). We mount small, circular active markers on the RGB-D camera and the objects for pose measurement. Examples of these markers are shown on the PR2's Kinect1 (b) and on the Kinect2 (c)}
		\label{fig:mocap-setup}
	\end{figure}

	%Deep neural networks require numerous data instances to support the millions of decision parameters. Moreover, in the case of semantic segmentation, we need pixel-level annotations on each RGB image in the dataset.
	
\begin{figure}
		\centering
		\subfigure{
			\includegraphics[width=0.2325\textwidth]{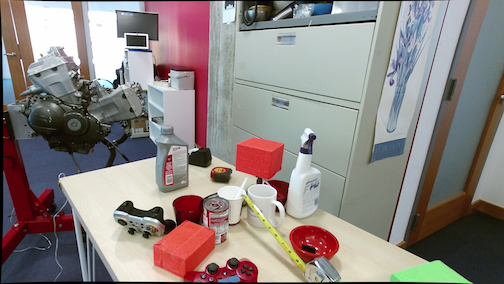}
			\includegraphics[width=0.2325\textwidth]{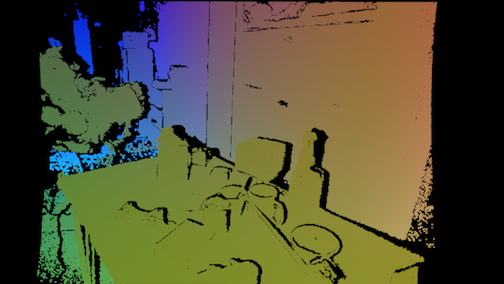}}%
		\vspace{-7pt}
		\subfigure{
			\includegraphics[width=0.2325\textwidth]{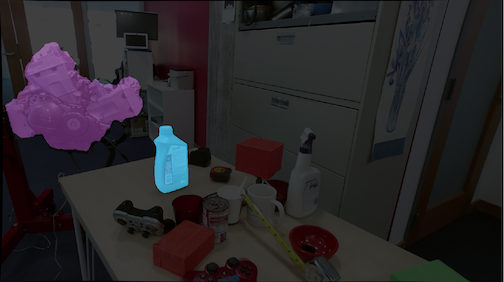}
			\includegraphics[width=0.2325\textwidth]{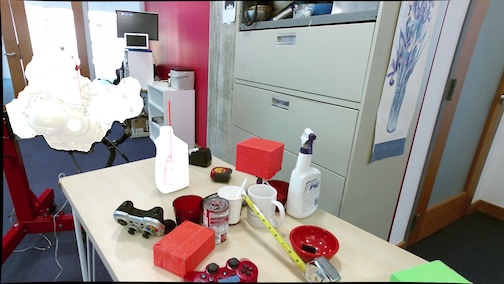}}%
		\vspace{-7pt}
		\caption{\textbf{Automatic Motion Capture Annotation:} Given input RGB and encoded depth images (top row), the automatic labeling system outputs the segmentation and object poses in axis angle format (bottom row).}
		\label{fig:mocap}
	\end{figure}

	We trained SegNet on $7500$ labeled images of indoor scenes consisting of automotive entities (e.g. engines, oil bottles, funnels, etc). Of these images, about two-thirds were hand-labeled by humans (using LabelMe \cite{russell2008labelme}) while the remaining third was generated automatically by a 3D Investigator\textsuperscript{TM} Motion Capture (MoCap) System and active markers placed on our cameras and objects (shown in Figure~\ref{fig:mocap-setup}). The training images span multiple sensor hardware (Microsoft Kinect1, Asus Xtion Pro Live, Microsoft Kinect2, and Carnegie Robotics Multisense SL) each with varying resolutions (respectively,  640$\times$480, 640$\times$480, 1280$\times$1024, and 960$\times$540). However, obtaining large datasets for segmentation and pose is difficult. As a result, we present a motion capture system to automatically annotate images shown in Figure~\ref{fig:mocap}. %, where input RGB and depth images are encoded into PNG format. The depth image is transformed into RGB using a 16-bit encoding, with each pixel's RGB channel values corresponding to pixel $x$, $y$, and $z$ in camera coordinates.% 
	Active markers are placed on the engine stand and on the corner of the table. Known transformations via MoCap are then used to segment the image by projecting a scanned object mesh using the transform into the camera optical frame, thus generating annotated segmentation and object pose data. 
	
	%	\begin{figure}
	%		\centering
	%		%\hspace{-.75cm}
	%		\includegraphics[width=0.485\textwidth]{images/dataset.png}
	%		\caption{\textbf{Draper Oil Change Dataset:} selected images consisting of various poses of objects from both Microsoft Kinect1 and Kinect2.}\label{dataset_sample}
	%		\label{dataset}
	%	\end{figure}
	
	\section{Evaluation}\label{sec:experiments}
	We benchmark SegICP on a dataset consisting of $1246$ annotated object poses obtained via the MoCap system. %; the dataset contains semantic labels of known objects and their poses relative to the camera. 
	%We benchmark SegICP on a dataset collected using the MoCap system in which we annotate the pose and segmentation using active markers on our sensor and objects. As described in Section~\ref{sec:data-collection}, the benchmark dataset consists of $1246$ annotated object poses generated through MoCap. 
	
	%; a subset of these are shown in Figure~\ref{dataset_sample} for both the Microsoft Kinect1 and Kinect2. %The distribution of annotated poses for selected objects (e.g. oil bottle and engine) are illustrated in Figure~\ref{dataset_histogram} as acquired from the MoCap system. 
	
	\begin{figure}
		\centering
		\subfigure{
			\includegraphics[height=0.0825\textheight, width=0.15\textwidth]{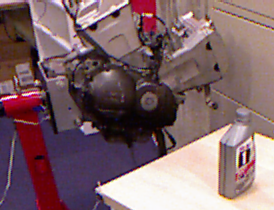}
			\includegraphics[height=0.0825\textheight, width=0.15\textwidth]{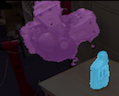}
			\includegraphics[height=0.0825\textheight, width=0.15\textwidth]{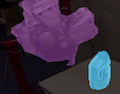}}%
		\vspace{-7pt}	
		\caption{\textbf{SegNet and DilatedNet:} outputs respectively (middle, right) given the same input RGB input image (left) from the PR2's Kinect1; SegNet appears to generate tighter segmentation compared to DilatedNet. }%
		\vspace{-15pt}
		\label{segnet-vs-dilated}
	\end{figure}
	
	\begin{figure*}
		\centering
		\includegraphics[width=\textwidth]{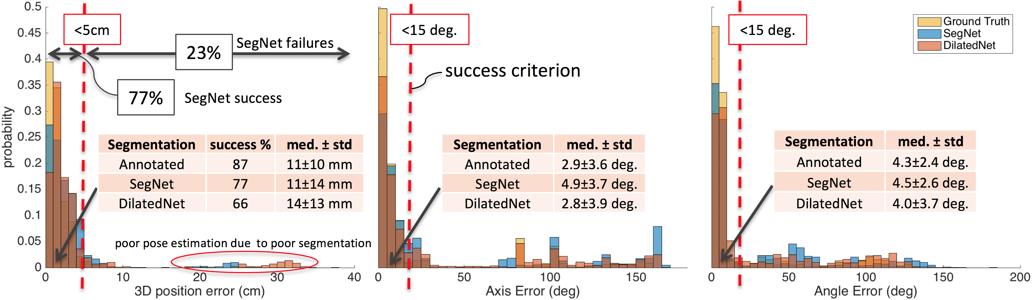}
		\caption{\textbf{SegICP Pose Estimation:} errors between varying segmentation masks (where Annotated refers to the annotated segmentation derived from MoCap) as a result of different neural network architectures (e.g. SegNet and DilatedNet).}
		\label{benchmarking}
	\end{figure*}	
	
	\subsection{Semantic Segmentation Results}
	\label{compare-segs}
	To categorize the influence of segmentation on pose estimation, we explored two architectures for semantic segmentation: SegNet and DilatedNet. SegNet is a computationally efficient autoencoder-decoder for pixel-wise semantic segmentation. The autoencoder architecture is essential for per-pixel classification, as it enables reconstruction of the inputs from the outputs at each layer, learning how to reconstruct the input before the final classification layer. DilatedNet makes use of dilated convolution modules to aggregate multi-scale contextual information without losing accuracy. Both network architectures adapt the convolutional layers of the VGG \cite{Simonyan2014} image classification network, with SegNet using the VGG layers as its encoder and DilatedNet converting later layers into dilated convolution modules. Weights are initialized during training using a VGG-16 model pretrained on ImageNet \cite{deng2009imagenet}. We train both of these networks with a dataset of over $7500$ annotated images (average epoch time of about an hour) and obtained the performance measures listed in Table~\ref{segmentation_quantitative}.

	\vspace{-2pt}
	\begin{table}[H]
		\centering
		\begin{tabular}{
				>{\columncolor[HTML]{C0C0C0}}c|c|c|}
			\cline{2-3}
			\multicolumn{1}{l|}{\cellcolor[HTML]{FFFFFF}}           & \multicolumn{1}{c|}{\cellcolor[HTML]{C0C0C0}\textbf{SegNet}} & \multicolumn{1}{c|}{\cellcolor[HTML]{C0C0C0}\textbf{DilatedNet}} \\ \hline
			\multicolumn{1}{|c|}{\cellcolor[HTML]{C0C0C0}\textbf{IOU median ($\pm$std)}}       & $0.850 \pm 0.159           $                                              & $0.752 \pm 0.189        $                                                      \\ \hline
			
			\multicolumn{1}{|l|}{\cellcolor[HTML]{C0C0C0}\textbf{Precision median ($\pm$std)}} & $0.897          \pm 0.107                                 $                 &$ 0.807               \pm0.183        $                                      \\ \hline
			
			\multicolumn{1}{|l|}{\cellcolor[HTML]{C0C0C0}\textbf{Recall median ($\pm$std)}}    & $0.961            \pm 0.164                             $                 & $0.965                 \pm  0.162       $                                        \\ \hline
			
		\end{tabular}
		\vspace{3pt}
		\caption{The performance of the semantic segmentation networks.}\label{segmentation_quantitative}
	\end{table}
	\vspace{-12pt}

	A key distinction between the two architectures is that DilatedNet was designed for increased recall by incorporating dilated convolution modules whereas SegNet appears to achieve higher precision measures. Notable visual differences are illustrated in Figure~\ref{segnet-vs-dilated}, where the output of SegNet and DilatedNet is displayed for the same scene.  %We found that SegNet exhibited higher IOU and precision measures as compared to DillatedNet, however, the latter architecture appeared to obtain slightly higher recall. 
	It is important to note that the \emph{quality} of the segmentation influences the point cloud mask and has immediate impact on the performance of our point-to-pose registration pipeline for object pose estimation. Still, the following questions still remain: \emph{Does higher segmentation IOU result in better pose? Higher precision? Higher recall?} In the next section, we perform several benchmarks to investigate these very questions.	
	
	\subsection{Pose Estimation Results}
	%We evaluate SegICP on the benchmark dataset generated from a motion capture system; the dataset contains semantic labels of known objects in the scene in addition to their poses relative to the camera. 
	
	\textbf{The Acquisition and Tracking Phases.}\;\; In our benchmarking, we used a collection of $N=30$ model crops for each object during the acquisition phase and discovered an overall average runtime of $270~\text{ms}$ over a collection of thirty threads on a six-core i7-6850K. However, note that the time evaluation here is directly dependent on the number of crops and the machine's CPU. The registration of each of these crops proposed a separate object pose hypothesis (alike Figure~\ref{multi-examples}), and we used a threshold of $\epsilon=0.75$ to switch into the tracking phase, which continuously updates the object's pose using the optimal crop, operating at about $70~\text{ms}$, with $45$--$50~\text{ms}$ being the neural network forward propagation (with nVidia GTX Titan X). For the kd-tree radius search to compute the metric, we used $\tau = 1~\text{cm}$. 
	
	\textbf{Benchmarking.}\;\; In Figure~\ref{benchmarking}, we illustrate the results of evaluating SegICP on the benchmarking dataset of $1246$ object pose annotations. To fully categorize the influence of the segmented mask on the final pose estimation, we ran SegICP using the annotated segmentation and the output of the two segmentation neural network architectures: SegNet and DilatedNet. These results indicate that SegNet achieves higher performance ($77\%$) as compared to DilatedNet ($66\%$). We categorize failure as exceeding errors of more than $5~\text{cm}$ in position and $15^\circ$ in axis and axis angle. These failures due to segmentation errors and model crop coverage represent a class of highlighted instances in the figure. Of the successful scenes, SegICP achieves $1~\text{cm}$ position error and $<5^\circ$ angle error; this level of accuracy corresponds to about $80\%$ of all the benchmarked instances. Further performance measures are given in Figure~\ref{benchmarking}, where we show the distribution of pose estimation errors given segmentation.
	%is as follows, 
	% \vspace{-5pt}
	% \begin{table}[H]
	% \centering
	% \begin{tabular}{l|c|c|c|}
	% \cline{2-4}                                                                & \cellcolor[HTML]{C0C0C0}\textbf{Annotated} & \cellcolor[HTML]{C0C0C0}\textbf{SegNet} & \cellcolor[HTML]{C0C0C0}\textbf{DilatedNet} \\ \hline
	
	% \multicolumn{1}{|l|}{\cellcolor[HTML]{C0C0C0}{\textbf{\%success}}} & $0.87               $                     & $0.77$                                &  $0.66$                            \\ \hline
	
	% \multicolumn{1}{|l|}{\cellcolor[HTML]{C0C0C0}{\textbf{pos err. $\pm$std (mm)}}} & $11               \pm 10$                              & $11      \pm 14$                                & $14    \pm 13$                                 \\ \hline
	
	% \multicolumn{1}{|l|}{\cellcolor[HTML]{C0C0C0}\textbf{axis err. $\pm$std (deg)}}        &      $   2.9    \pm   3.6$      &                                          $    4.9    \pm 3.7$                         & $2.8      \pm 3.9$                                  \\ \hline
	% \multicolumn{1}{|l|}{\cellcolor[HTML]{C0C0C0}\textbf{angle err. $\pm$std (deg)}}                          & $4.3 \pm2.4     $                                           & $4.5      \pm 2.6   $                               &$ 4.0    \pm 3.7 $                                     \\ \hline
	% \end{tabular}
	% \vspace{3pt}
	% \caption{\textbf{Pose estimation performance:} A comparison between SegNet and DilatedNet segmentation masks for pose estimation}
	% \label{pose-errors}
	% \end{table}
	% \vspace{-15pt}
	
	Interestingly, the performance of SegICP is highly correlated with both sensor technology and calibration. When considering only the $466$ Kinect1 instances (a structured light sensor with better RGB-D calibration), SegICP achieves success measures of $90\%$, $73\%$, and $72\%$ using segmented masks from annotation, SegNet, and DilatedNet respectively; the networks appear to have comparable performance. However, when calibration is subpar, in the case of our Kinect2 (which is also a time of flight sensor), it is beneficial to bound the number of false-positive pixels (maximizing precision) to avoid acquiring misaligned points in the cropped scene cloud. From Table~\ref{segmentation_quantitative}, SegNet and DilatedNet achieve precision measures of $0.897$ and $0.807$ respectively.  With the Kinect2, we observe success measures of $85\%$, $80\%$, and $62\%$ for annotated, SegNet, and DilatedNet segmentation; the large inconsistencies with DilatedNet is as a result of poor cropped scene clouds due to excessive false-positives in the segmentation (poor precision). 
	
\begin{figure}
		\includegraphics[width=0.475\textwidth]{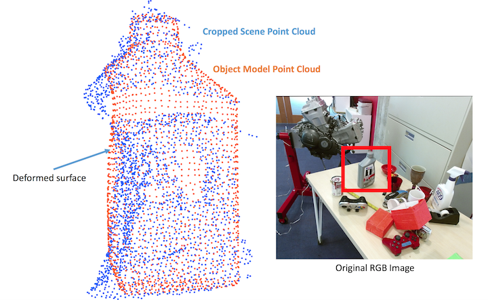}
		\caption{\textbf{Point Cloud Deformation:} experienced by time of flight sensors (e.g. Microsoft Kinect2) on reflective surfaces and a culprit of object model cloud misalignments in the dataset.} \label{fig:deform}
	\end{figure}

	Further, it appears that SegICP operates with higher performance on structured light sensors (e.g. Kinect1) compared to time of flight sensors (e.g. Kinect2). We discovered that objects with reflective surfaces (e.g. the oil bottle) with high levels of geometric symmetry and potential ambiguities result in poor ICP fits due to the deformations in the point cloud caused by time of flight. Figure~\ref{fig:deform} illustrates this particular phenomenon, where large deformities on the surface of the oil bottle is present, resulting in poor registration. Lastly, because the architecture uses a segmentation mask (generated using the RGB frames) to crop the point cloud, the sensor calibration of the RGB and depth frames is crucial for accurate pose estimation. 
	
	\section{Conclusion}    
	We present a novel, \emph{highly parallelized} architecture for semantic segmentation and accurate pose estimation ($1~\text{cm}$ position error and ${<}\,5^\circ$ angle error). Our architecture delivers immediate benefits as compared to work in the literature by not requiring an initial guess sufficiently close to the solution and by being inherently parallelizable, allowing us to process multiple objects simultaneously in real time ($70$--$270~\text{ms}$ in tracking and acquisition mode respectively).  We elaborated on a motion capture approach to collecting potentially massive sets of annotated segmentation and pose data, allowing our architecture to scale rapidly to more enriched domains. Lastly, we categorized the segmentation-driven method to pose estimation by extensively investigating and benchmarking two different neural network architectures. 
	
	We are currently working to refine the perception architecture, extend the framework to incorporate much larger sets of objects and tie it with integrated task and motion planning for complex interactions in unstructured environments.
	
	\bibliographystyle{IEEEtran}
	\footnotesize{\bibliography{belief,deep-learning,icp,segnet}}
\end{document}